\documentclass[journal]{IEEEtran}
\usepackage{ifpdf}

\ifCLASSOPTIONcompsoc
  \usepackage[nocompress]{cite}
\else
  \usepackage{cite}
\fi

\ifCLASSINFOpdf
  \usepackage[pdftex]{graphicx}
\else
  \usepackage[dvips]{graphicx}
\fi

\usepackage{amsmath}
\usepackage[linesnumbered,ruled]{algorithm2e}
\usepackage{array}
\usepackage{mdwmath}
\usepackage{mdwtab}
\usepackage{multirow}
\usepackage[caption=false,font=footnotesize,labelfont=sf,textfont=sf]{subfig}

\usepackage{url}

\begin{document}
%
\title{Automatically Designing CNN Architectures Using Genetic Algorithm for Image Classification}
%
%
%
%

\author{Yanan~Sun,~\IEEEmembership{Member,~IEEE,}
        ~Bing~Xue,~\IEEEmembership{Member,~IEEE,}
        ~Mengjie~Zhang,~\IEEEmembership{Fellow,~IEEE,}
         Gary~G.~Yen,~\IEEEmembership{Fellow,~IEEE,} 
         and Jiancheng Lv,~\IEEEmembership{Member,~IEEE,}
\thanks{This paper has been accepted by IEEE Transactions on Cybernetics, DOI:10.1109/TCYB.2020.2983860}

}

\IEEEtitleabstractindextext{%
\begin{abstract}
Convolutional Neural Networks (CNNs) have gained a remarkable success on many image classification tasks in recent years. However, the performance of CNNs highly relies upon their architectures. For most state-of-the-art CNNs, their architectures are often manually-designed with expertise in both CNNs and the investigated problems. Therefore, it is difficult for users, who have no extended expertise in CNNs, to design optimal CNN architectures for their own image classification problems of interest. In this paper, we propose an automatic CNN architecture design method by using genetic algorithms, to effectively address the image classification tasks. The most merit of the proposed algorithm remains in its ``automatic'' characteristic that users do not need domain knowledge of CNNs when using the proposed algorithm, while they can still obtain a promising CNN architecture for the given images. The proposed algorithm is validated on widely used benchmark image classification datasets, by comparing to the state-of-the-art peer competitors covering eight manually-designed CNNs, seven automatic+manually tuning and five automatic CNN architecture design algorithms. The experimental results indicate the proposed algorithm outperforms the existing automatic CNN architecture design algorithms in terms of classification accuracy, parameter numbers and consumed computational resources. The proposed algorithm also shows the very comparable classification accuracy to the best one from manually-designed and automatic+manually tuning CNNs, while consumes much less of computational resource.
\end{abstract}

}

\maketitle

\IEEEdisplaynontitleabstractindextext

%
\IEEEpeerreviewmaketitle

\section{Introduction}
\label{section_introduction}

\IEEEPARstart{C}{onvolutional} Neural Networks (CNNs), as the dominant technique of deep learning~\cite{lecun2015deep}, have shown remarkable superiority in various real-world applications over most machine learning approaches~\cite{krizhevsky2012imagenet,sainath2013deep,sutskever2014sequence,alphago}. It has been known that the performance of CNNs highly relies upon their architectures~\cite{krizhevsky2012imagenet,simonyan2014very}. To achieve promising performance, the architectures of state-of-the-art CNNs, such as GoogleNet~\cite{szegedy2015going}, ResNet~\cite{he2016deep} and DenseNet~\cite{huang2017densely}, are all manually designed by experts who have rich domain knowledge from both investigated data and CNNs. Unfortunately, such domain knowledge is not necessarily held by each interested user. For example, users who are familiar with the data at hand do not necessarily have the experience in designing the architectures of CNNs, and vice versa. As a result, there is a surge of interest in automating the design of CNN architectures, allowing tuning skills of the CNN architectures to be transparent to the users who have no domain knowledge on CNNs. On the other hand, CNN architecture design algorithms can also spread the wide adoption of CNNs, which in turn promotes the development of machine intelligence.

Existing CNN architecture design algorithms can be divided into two different categories, based on whether domain knowledge is required or not when using them. The first is the ``automatic + manually tuning" CNN architecture designs, which implies that the manual tuning based on the expertise in designing CNN architectures is still required. This category covers the genetic CNN method (Genetic CNN)~\cite{xie2017genetic}, the hierarchical representation method (Hierarchical Evolution)~\cite{liu2017hierarchical}, the efficient architecture search method (EAS)~\cite{cai2018efficient}, the block design method (Block-QNN-S)~\cite{zhong2017practical}, and the advanced neural architecture search method (NSANet)~\cite{zoph2018learning}. The other is the ``automatic" CNN architecture designs, which do not require any manually-tuning from users when using them. The large-scale evolution method (Large-scale Evolution)~\cite{real2017large}, the Cartesian genetic programming method (CGP-CNN)~\cite{suganuma2017genetic}, the neural architecture search method (NAS)~\cite{zoph2016neural}, and the meta-modelling method (MetaQNN)~\cite{baker2016designing} belong to this category. Owing to the extra benefits of manual expertise in CNNs, it is natural that the ``automatic+manually tuning'' designs often show slightly better performance than the ``automatic'' designs. However, the significant superiority of the ``automatic'' designs is their manually-tuning-free characteristic, which is much preferred by users without any domain knowledge of CNNs. For example, the NASNet and the Large-scale Evolution algorithms, which are from the ``automatic+manually tuning'' and the ``automatic'' categories, respectively, achieved the classification accuracies of 96.27\% and 94.60\% on the CIFAR10 benchmark dataset~\cite{krizhevsky2009learning}, respectively. However, a CNN framework architecture must be provided to NSANet because NSANet only automated partial cells of the framework. Obviously, if the framework architecture is not well-designed, the resulted CNN will not have promising performance. When using the Large-scale Evolution method, the users just directly perform it on the given data, and finally a promising CNN architecture is obtained. To this end, the ``automatic'' CNN architecture designs should be more welcomed because the majority of CNN users have no extensive domain knowledge of CNN architecture design.

On the other hand, based on the adopted techniques, CNN architecture designs can also be classified into the evolutionary algorithm-based ones and the reinforcement learning-based ones. Specifically, Genetic CNN, Large-scale Evolution, Hierarchical Evolution and CGP-CNN are based on evolutionary algorithms~\cite{back1996evolutionary}, following the standard flow of an evolutionary algorithm to heuristically discover the optimal solution; while NAS, MetaQNN, EAS, Block-QNN-S and NASNet are based on reinforcement learning~\cite{sutton1998reinforcement}, resembling those based on evolutionary algorithms, in addition to the employed heuristic nature utilizing the reward-penalty principle of reinforcement learning. Experimental evidence shows that the reinforcement learning-based designs often require more intensive computational resources than the ones based on the evolutionary algorithms~\cite{sun2017evolving}. For example, the NAS method consumed 800 Graphic Processing Units (GPUs) in 28 days to find the promising CNN architecture on the CIFAR10 dataset, while the Genetic CNN consumed only 17 GPUs in one day on the same dataset providing with similar performance. To this end, the evolutionary algorithms-based CNN architecture designs are much preferred because intensive computational resources are not necessarily available to every interested user.

Evolutionary algorithm~\cite{back1996evolutionary} is a class of population-based meta-heuristic optimization paradigms inspired by the biological evolution. Typical evolutionary algorithms include genetic algorithms (GAs)~\cite{davis1991handbook}, genetic programming~\cite{banzhaf1998genetic}, evolutionary strategy~\cite{janis1976evolutionary}, etc., among which GAs are the most popular one mainly because of their theoretical evidences~\cite{schmitt2001theory} and promising performance in solving different optimization problems~\cite{sun2018igd,deb2002fast,sun2017reference,transferjiang2018,sun2018improve}. It has also been recognized that GAs are capable of generating high-quality optimal solutions by using bio-inspired operators, i.e., mutation, crossover and selection~\cite{mitchell1998introduction}. Commonly, the operators need to be carefully designed for the specific problems at hand. For example, Liu \textit{et al.}~\cite{LiuOn} designed a one-point crossover operator and a novel mutation operator for the WCDMA network planning. In addition, Kaustuv \textit{et al.}~\cite{Nag2015A} developed a new mutation operator to exploit the fitness and the unfitness for feature selection via an ensemble method for feature selection and classification. Considering both characteristics of ``automatic'' and evolutionary algorithm-based CNN architecture designs, we propose an effective and efficient algorithm by using GA, in short, termed as CNN-GA, to automatically discover the best architectures of CNNs, so that the discovered CNN can be directly used without any manual tunings. Obviously, CNN-GA is an automatic CNN architecture design algorithm. Please note that the ``automatic'' and ``automatic+manually-tuning'' are discussed from the users' view but not the developers' view. In contrast, adequate domain knowledge should be encouraged in developing high-performance CNN architecture design algorithms. This effort can be easily understood by the analogy to the design of Windows Operation System by scientists from Microsoft: the scientists should use their professional knowledge as much as possible in designing a user-friendly operating system so that the users are allowed to effectively work on the computers, even without extensive knowledge on the operating system.

The novelty of the proposed algorithm lies in its complete automation in discovering the best CNN architectures, and not requiring any manual intervention during the evolutionary search. The contributions of the proposed CNN-GA algorithm are summarized as follows:

\begin{enumerate}
	\item GAs often employ the fixed-length encoding strategy because of the crossover operators primitively designed for the individuals having the same lengths. In this case, the length of the encoding must be specified beforehand. Ideally, the length should be the optimal CNN depth that is mostly unknown in advance. As a result, the specified number may be incorrectly estimated, resulting in ineffective architecture designed. Although many researchers have developed the variable-length encoding strategy independently, the resulted CNN architecture is not optimal because the crossover operator is not redesigned accordingly. In this paper, we have proposed a variable-length encoding strategy and the corresponding crossover operator to address both issues aforementioned efficiently and effectively.

	\item Most existing CNN architecture algorithms are designed based on the basic components or the well-constructed blocks of CNNs. In practice, both designs often generate the ineffective CNN architectures and the complex CNN architectures having poor generalization ability, respectively. In this paper, the skip connections are incorporated into the proposed algorithm to deal with complex data by avoiding the Vanishing Gradient (VG) problems~\cite{hochreiter1997long}. On one hand, this design can reduce the search space so that the best performance can be easily identified. On the other hand, compared to other algorithms with similar performance, the architectures evolved by the proposed algorithm are much simpler.

	\item In order to speed up the CNN architecture design and provide the users with an optimal CNN architecture within an acceptable time, most of the existing algorithms employed extensive computational resource. Particularly, the algorithms employed the data parallelism strategy which is not efficient. The main reason is that most architectures are median-scale and do not need to run on the extensive computational resource. Otherwise, the communication consumption between different computational nodes will take up most of the computational cost. In the proposed algorithm, an asynchronous computational component is developed to make full use of the given computational resources to accelerate the evaluation of the fitness of the individuals, while a cache component is employed to further reduce the fitness evaluation time for the whole population.

\end{enumerate}

The remainder of this paper is organized as follows. Firstly, related works and background are presented in Section~\ref{section_literature}. Then, the details of the proposed algorithm are documented in Section~\ref{section_algorithm}. Next, the experiment designs and experimental results as well as the analysis are shown in Sections~\ref{section_exp_settings} and~\ref{section_exp_results}, respectively. Finally, conclusions and future works are outlined in Section~\ref{section_conclusion}.

\section{Literature Review}
\label{section_literature}

In this section, CNNs and skip connections which are considered the background of the proposed algorithm, are introduced to help readers better understand the related works and the proposed algorithm. Then, the relevant work in discovering the architectures of CNNs is reviewed.

\subsection{Background}
\label{sub_section_background}
\subsubsection{Convolutional Neural Networks}
\label{sub_sub_section_cnn} 

In this subsection, we mainly introduce the building blocks of CNNs, i.e., the convolutional and pooling layers, which are the basic objects encoded by GAs to represent CNNs.

Specifically, the convolutional layer employs filters to perform convolutional operations on the input data. One filter can be viewed as a matrix. During the convolutional operation, the filter horizontally slides (with a given step size), then vertically moves (with another step size) for the next horizontal slide, until the whole image has been scanned. The set of filter outputs form a new matrix called the feature map. The horizontal and vertical step sizes are called the width and height of a stride. The exact number of feature maps used is a parameter in the architecture of the corresponding CNN. In addition, two convolutional operations are applied: the \textit{same} convolutional operation which pads zeros to the input data when there is not enough area for the
filter to overlap, and the \textit{valid} convolutional operation which does not pad anything. Hence, the parameters of a convolutional layer are the number of feature maps, the filter size, the stride size and the convolutional operation type. A pooling layer has common components of a convolutional layer except that 1) the filter is called the kernel which has no value, 2) the output of a kernel is the maximal or mean value of the area it stops, and 3) the spatial size of the input data is not changed through a pooling layer. When the maximal value is returned, it is a pooling layer with the type of \textit{max}, otherwise of \textit{mean}. Hence, the parameters of a pooling layer are the kernel size, the stride size and the pooling type used. In addition, the fully-connected layers are usually incorporated into the tail of a CNN. In the proposed algorithm, the fully-connected layers are discarded, and the justifications are given in Subsection~\ref{sub_section_pop_init}.

\subsubsection{Skip Connections}
\label{sub_sub_section_skip_connections}
\begin{figure}[!htp]
	\centering
	\includegraphics[width=\columnwidth]{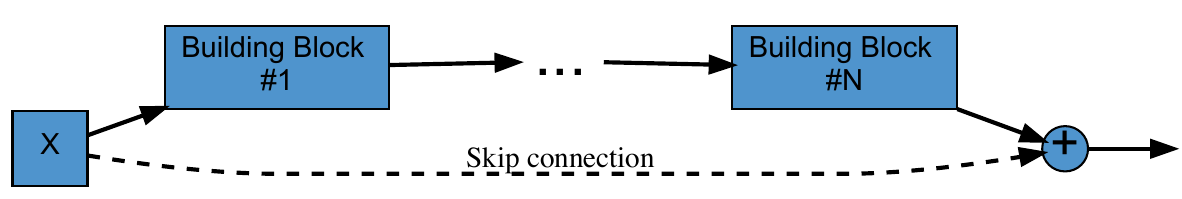}\\
	\caption{An example of using the skip connection.}\label{fig_skip_connection}
\end{figure}

The skip connections refer to those connecting the neurons of the layers that are not adjacent. The skip connection was firstly introduced in~\cite{hochreiter1997long} as a gate mechanism, effectively training a recurrent neural network with long and short-term memory~\cite{gers1999learning} and avoiding the VG problems~\cite{hochreiter1997long}. Specifically, the VG problems refer to the gradient becoming either very small or explosion during back propagation training in a deep neural network, and are the main obstacle to effectively train \textit{deep} neural networks. The skip connections were experimentally proven to be able to train very deep neural networks~\cite{srivastava2015training}. Indeed, the promising performance of ResNet~\cite{he2016deep}, which was proposed very recently, also benefits from the skip connections. A simple example of using a skip connection is shown in Fig.~\ref{fig_skip_connection}, where the dashed line denotes the skip connection from the input $X$ to the output of the $N$-th building block, and the symbol ``$\oplus$'' refers to the element-wise addition.

Over the past few years, an increasing number of researchers has attempted to theoretically reveal the mechanisms behind the skip connections. For example, the skip connections have also been claimed to be able to eliminate singularities~\cite{orhan2017skip}. Among the existing theoretical evidence, skip connections defying the VG problems receives the most recognition~\cite{gers1999learning,hochreiter1997long}. Because the skip connections shorten the number of layers of back-propagation, the VG problems should be alleviated. As discussed above, the deeper the CNN is, the more powerful capability it would have to process complex data. Combined with the connections that are not skipped, a CNN with the skip connections can have the capability that a deep architecture has and can also be effectively trained.

\begin{figure*}[!htp]
	\centering
	\includegraphics[width=1.8\columnwidth]{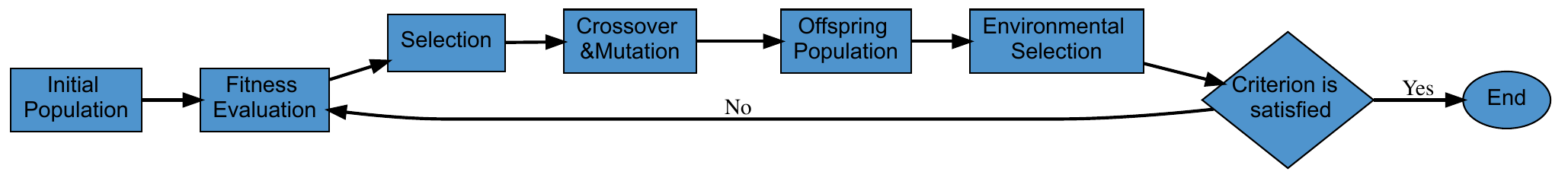}
	\caption{The flowchart of a genetic algorithm.}\label{fig_ga_flow}
\end{figure*}


\subsection{Related Work}
\label{sub_section_realted_works}
As the based work of the proposed algorithm, the Genetic CNN method~\cite{xie2017genetic} is discussed in this subsection. 

In Genetic CNN, encoding the whole architecture of a CNN is composed of multiple stages. In each stage, a small architecture is encoded, and then multiple, different small architectures are stacked to give the whole architecture. Just like the NSANet method, a CNN architecture framework is provided to Genetic CNN, while the small architectures replace the convolutional layers in the provided framework to form the final CNN. These small architectures are called cell in Genetic CNN, and each cell is designed in each stage of Genetic CNN. In each stage, multiple predefined building blocks of CNNs are ordered, and their connections are encoded. For ordering these building blocks, the first and the last building blocks are manually specified, and the remainings are all convolutional layers with the same settings. In order to encode the connections between these ordered building blocks, a binary string encoding method~\cite{srinivas1994genetic} is used. Suppose there are four building blocks, a string of ``111-10-0'' implies that the first building block has the connections to all other building blocks. The second one has no connection to the fourth building block, which is the same as the third one. Furthermore, the number of stages is manually predefined. Clearly observed from this encoding strategy, multiple manual interventions are required, such as the numbers of building blocks in each stage and total stages. Because these numbers are related to the depth of the discovered CNN, domain expertise in CNNs is strongly needed to predefine these numbers for proper depth of the CNN. In addition, the parameters of the building blocks in each stage cannot be changed in Genetic CNN. Indeed, this encoding strategy can be viewed at only discovering the connections of a given CNN, and cannot change the depth of the given CNN. Note that, no speeding up technique was designed in Genetic CNN, resulting in the experiments not applicable to complex datasets such as CIFAR100.

\begin{figure*}[!htp]
	\centering
	\includegraphics[width=2\columnwidth]{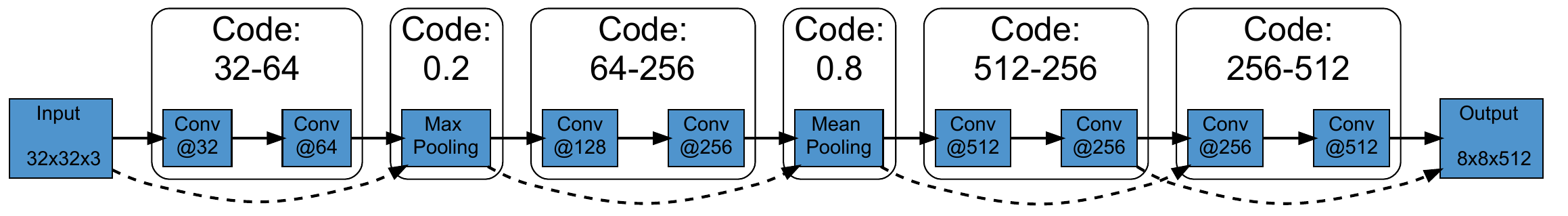}\\
	\caption{An example of the proposed encoding strategy representing a CNN. This CNN is composed of four skip layers and two pooling layers. The code encoding this CNN is composed of the codes representing each layer. For each skip layer, its codes are the number of feature maps of the two convolutional layers within this skip layer; while the code for each pooling layer is the pooling type. We use a number between $(0, 0.5)$ to represent the max pooling type, and a number between $[0.5, 1)$ to represent the mean pooling type. The numbers above each layer are the code of the corresponding layer. So the code representing this CNN is ``32-64-0.2-64-256-0.8-512-256-256-512''.}\label{fig_overall_encoding}
\end{figure*}

\section{The Proposed Algorithm}
\label{section_algorithm}

In this section, we firstly present the framework of the proposed algorithm in Subsection~\ref{sub_section_alg_overview}, and then detail the main steps in Subsections~\ref{sub_section_pop_init} to \ref{sub_section_environmental_selection}. To help readers better understand the proposed algorithm, we will not only document the details of each main step, but also provide the analysis for given designs.

\subsection{Algorithm Overview}
\label{sub_section_alg_overview}
\begin{algorithm}
	\label{alg_framework}
	\caption{The Proposed Algorithm}
	\KwIn{A set of predefined building blocks, the population size, the maximal generation number, the image dataset for classification.}
	\KwOut{The discovered best architecture of CNN.}
	$P_0\leftarrow$ Initialize a population with the given population size using \textbf{the proposed variable-length encoding strategy}\;
	\label{alg_framework_line1}
	$t\leftarrow 0$\;
	\label{alg_framework_line2}
	\While{$t<$ the maximal generation number}
	{
		Evaluate the fitness of each individual in $P_t$ using \textbf{the proposed acceleration components}\;
		\label{alg_framework_evo_line1}
		$Q_t\leftarrow$ Generate offspring from the selected parent individuals using \textbf{the proposed mutation and the crossover operators}\;
		\label{alg_framework_evo_line2}
		$P_{t+1}\leftarrow$ Environmental selection from $P_t\cup Q_t$\;
		\label{alg_framework_evo_line3}
		$t\leftarrow t+1$\;
	}
	\textbf{Return} the individual having the best fitness in $P_{t}$.
\end{algorithm}
Algorithm~\ref{alg_framework} shows the framework of the proposed algorithm. Specifically, by giving a set of predefined building blocks of CNNs, the population size as well as the maximal generation number for the GA and the image classification dataset, the proposed algorithm begins to work, through a series of evolutionary processes, and finally discovers the best architecture of the CNN to classify the given image dataset. During evolution, a population is randomly initialized with the predefined population size, using the proposed encoding strategy to encode the predefined building blocks (line~\ref{alg_framework_line1}). Then, a counter for the current generation is initialized to zero (line~\ref{alg_framework_line2}). During evolution, the fitness of each individual, which encodes a particular architecture of the CNN, is evaluated on the given dataset (line~\ref{alg_framework_evo_line1}). After that, the parent individuals are selected based on the fitness, and then generate new offspring by the genetic operators including the crossover and mutation operators (line~\ref{alg_framework_evo_line2}). Then, a population of individuals surviving into the next generation is selected by the environmental selection from the current population (line~\ref{alg_framework_evo_line3}). Specifically, the current population is composed of the parent population and the generated offspring population. Finally, the counter is increased by one, and the evolution continues until the counter exceeds the predefined maximal generation. As shown in Fig.~\ref{fig_ga_flow}, the proposed algorithm follows the standard pipeline of a GA (the phases of selection, crossover and mutation, and the offspring population shown in Fig.~\ref{fig_ga_flow} are collectively described in line~\ref{alg_framework_evo_line2} of Algorithm~\ref{alg_framework}).

Please note that GAs only provide a unified framework to solve optimization problems with their inherent biological mechanism. When GAs are used in practice, their components regarding the biological mechanism must be specifically designed based on the particular problems to be solved. In the proposed algorithm, we carefully design the variable-length encoding strategy, acceleration components and genetic operators (they are also highlighted in bold in Algorithm~\ref{alg_framework}), to assure the effectiveness and efficiency of the proposed algorithm in designing CNN architectures.

\subsection{Population Initialization}

\label{sub_section_pop_init}
As introduced in Section~\ref{section_literature}, a CNN is composed of the convolutional layers, pooling layers and occasionally fully-connected layers. The performance of a CNN highly relies on its depth, and the skip connections could turn the \textit{deep depth} to be a reality. In the proposed encoding strategy, we design a new building block by directly using the skip connections, named the skip layer, to replace the convolutional layer when forming a CNN. In addition, the fully-connected layers are discarded in the proposed encoding strategy (the reason will be given later in this subsection). In summary, only the skip layers and the pooling layers are used to construct a CNN in the proposed encoding strategy.

Specifically, a skip layer is composed of two convolutional layers and one skip connection. The skip connection connects from the input of the first convolutional layer to the output of the second convolutional layer. As introduced above, the parameters of a convolutional layer are the number of feature maps, the filter size, the stride size and the convolutional operation type. In the proposed encoding strategy, we use the same settings for the filter sizes, stride sizes and convolutional operations, respectively. Particularly, the filter and stride sizes are set to $3\times 3 $ and $1\times 1$, respectively, and only the \textit{same} convolutional operation is used. To this end, the parameters encoded for a skip layer are the numbers of the feature maps for the two convolutional layers (denoted as $F1$ and $F2$, respectively). In addition, the pooling layers used in the proposed encoding strategy are set to be $2\times 2$ for the kernel sizes and the stride sizes. To this end, the parameter encoded for a pooling layer is only the pooling type (denoted as $P1$). Note that, the reasons for this design and adopting the settings for such a design will be explained later in this subsection.

\begin{algorithm}
	\label{alg_pop_init}
	\caption{Population Initialization}
	\KwIn{The population size $T$.}
	\KwOut{The initialized population $P_0$.}
	$P_0\leftarrow \emptyset$ \;
	\While{$|P_0|<T$}
	{
		$L\leftarrow$ Randomly generate an integer greater than zero\;
		\label{alg_pop_init_line1}
		$list\leftarrow$ Create a linked list contains $L$ nodes\;
		\label{alg_pop_init_line2}
		\ForEach{node in the linked list}
		{\label{alg_pop_init_node_begin}
			$r\leftarrow$ Uniformly generate a number from $(0,1)$\;
			\label{alg_pop_init_node_type}
			\uIf{$r<0.5$}
			{\label{alg_pop_init_skip_layer_begin}
				$node.type\leftarrow 1$\;
				$node.F1\leftarrow Randomly$ generate an integer greater than zero\;
				$node.F2\leftarrow Randomly$ generate an integer greater than zero\;
			}\label{alg_pop_init_skip_layer_end}
			\Else{\label{alg_pop_init_pool_begin}
				$node.type\leftarrow 2$\;
				$q\leftarrow$ Uniformly generate a number from $(0,1)$\;
				\uIf{$q<0.5$}
				{
					$node.P1\leftarrow$ \textit{max}\;
				}
				\Else{
					$node.P1\leftarrow$ \textit{mean}\;	
				}
			}\label{alg_pop_init_pool_end}
		}	\label{alg_pop_init_node_end}
		$P_0\leftarrow P_0\cup list$\;
		\label{alg_pop_init_end}
	}
	\textbf{Return} $P_{0}$.

\end{algorithm}
Algorithm~\ref{alg_pop_init} shows the details of the population initialization. Briefly, $T$ individuals are initialized with the same ways, and then they are stored into $P_0$. During the individual initialization process, the length (denoted as $L$) of an individual, representing the depth of the corresponding CNN, is randomly initialized firstly (line~\ref{alg_pop_init_line1}). Then, a linked list containing $L$ nodes is created (line~\ref{alg_pop_init_line2}). After that, each node is configured (lines~\ref{alg_pop_init_node_begin}-\ref{alg_pop_init_node_end}), and then the linked list is stored into $P_0$ (line~\ref{alg_pop_init_end}). During the configuration of each node, a number, $r$, is randomly generated from $(0,1)$ (line~\ref{alg_pop_init_node_type}). If $r<0.5$, the type of this node is marked as a skip layer by setting its $type$ property to $1$. Otherwise, this node represents a pooling layer by setting $type$ to $2$. In the case of a skip connection layer, the numbers of the feature maps are randomly generated and then assigned to $node.F1$ and $node.F2$, respectively (lines~\ref{alg_pop_init_skip_layer_begin}-\ref{alg_pop_init_skip_layer_end}). Otherwise, the pooling type is determined by the probability of tolling a coin. Particularly, the pooling type, $node.P1$, is set to \textit{max} when the probability is below $0.5$, and \textit{mean} otherwise (lines~\ref{alg_pop_init_pool_begin}-\ref{alg_pop_init_pool_end}). 

An example of the proposed encoding strategy encoding a CNN is shown in Fig.~\ref{fig_overall_encoding}. This CNN is composed of four skip layers and two pooling layers. The code representing one skip layer is the string of the feature map numbers of the corresponding convolutional layers within the same skip layer; while that for a pooling layer is a number representing the pooling type. Particularly, a random number between $(0, 0.5)$ represents a max pooling layer, while that between $[0.5, 1)$ represents a mean pooling layer. The code representing the whole CNN is the sequential string connection of codes representing the layers. As shown in this example where the code of each layer is listed above itself, the code of the whole CNN is ``32-64-0.2-64-256-0.8-512-256-256-512'' to represent a CNN with a depth of $10$.

Next, we will detail the reasons why discarding the fully-connected layers, using two convolutional layers in a skip layer and the settings for the skip and pooling layers in the proposed algorithm. Specifically, multiple fully-connected layers are typically added to the tail of a CNN. However, the fully-connected layer easily results in the over-fitting phenomenon~\cite{hawkins2004problem} due to its dense connection~\cite{srivastava2014dropout}. To reduce the over-fitting, the dropout~\cite{srivastava2014dropout} randomly removing a part of the connections is commonly used. However, each dropout will introduce one parameter. Only a properly specified parameter can lead to the promising performance of the corresponding CNN. Meanwhile, the number of fully-connected layers and the number of neurons in each fully-connected layer are also two parameters hard to tune. If the fully-connected layers are incorporated into the proposed encoding strategy, the search space will substantially enlarge, increasing the difficulty of finding the best CNN architecture. The use of two convolutional layers in a skip layer is inspired by the design of ResNet, and the effectiveness of such skip layers have been experimentally proved in literature~\cite{huang2017densely,he2016identity,real2017large,liu2017hierarchical}. However, the sizes of feature maps in each skip layer of ResNet are set to be equal. In the proposed encoding strategy, the sizes of feature maps can be unequal, which is more flexible. Furthermore, setting the convolutional operation to \textit{same} and using the $1\times 1$ stride are to make the dimension of the input data remain the same, which is more flexible for such an automatic design because $1\times 1$ stride does not change the image size. As the settings of filter and kernel sizes as well as the stride size in the pooling layers, they are all based on the designs of existing hand-crafted CNNs~\cite{huang2017densely,he2016identity}. Moreover, another important reason for specifying such settings is based on our expertise in manually tuning the architectures of CNNs. The effectiveness of such settings will be shown in Section~\ref{section_exp_results}.

\subsection{Fitness Evaluation}

\begin{algorithm}
	\label{alg_fitness_eval}
	\caption{Fitness Evaluation}
	\KwIn{The population $P_t$ of the individuals to be evaluated, the image dataset for classification.}
	\KwOut{The population $P_t$ of the individuals with their fitness values.}
	\If{t == 0}
	{\label{alg_cache_begin}
		$Cache \leftarrow \emptyset$\;
		Set $Cache$ to a global variable\; 
		
	}\label{alg_cache_end}
	\ForEach{individual in $P_t$}
	{	
		\eIf{the identifier of individual in Cache}
		{\label{alg_cache_get_from}
			$v\leftarrow$ Query the fitness by $identifier$ from $Cache$\;
			Set $v$ to $individual$\;
			\label{alg_cache_get_end}
		}
		{\label{alg_cache_not_from}
			\While{there is available GPU}
			{
				asynchronously evaluate $individual$ in an available GPU (details shown in Algorithm~\ref{alg_fitness_individual_eval})\;
			}
		}\label{alg_cache_not_end}
		
	}
	\textbf{Return} $P_{t}$.

\end{algorithm}

Algorithm~\ref{alg_fitness_eval} details the fitness evaluation of the individuals in the population $P_t$. Briefly, given the population, $P_t$, containing all the individuals for evaluating the fitness, and the image classification dataset on which the best architecture of a CNN is to discover, Algorithm~\ref{alg_fitness_eval} evaluates each individual of $P_t$ in the same manner, and finally returns $P_t$ containing the individuals whose fitness have been evaluated. Specifically, if the fitness evaluation is for the initialized population, i.e., $P_0$, a global cache system (denoted as $Cache$) is created, storing the fitness of the individuals with unseen architectures (lines~\ref{alg_cache_begin}-\ref{alg_cache_end}). For each individual (denoted by $individual$) in $P_t$, if $individual$ is found in $Cache$, its fitness is directly gotten from $Cache$ (lines~\ref{alg_cache_get_from}-\ref{alg_cache_get_end}). Otherwise, $individual$ is asynchronously placed on an available GPU for its fitness evaluation (lines~\ref{alg_cache_not_from}-\ref{alg_cache_not_end}). Note that, querying an individual from $Cache$ is based on the individual's identifier. Theoretically, arbitrary identifiers can be used, as long as they can distinguish individuals encoding different architectures. In the proposed algorithm, the 224-hash code~\cite{housley2004224}, which has been implemented by most programming languages, in terms of the encoded architecture is used as the corresponding identifier. Furthermore, the individual is asynchronously placed on an available GPU, which implies that we don't need to wait for the fitness evaluation for the next individual until the fitness evaluation of the current one finishes, but place the next individual on an available GPU immediately.

\begin{algorithm}
	\label{alg_fitness_individual_eval}
	\caption{Individual Fitness Evaluation}
	\KwIn{The individual $individual$, the available GPU, the number of training epochs, the global cache $Cache$, the training data $D_{train}$ and the fitness evaluation data $D_{fitness}$ from the given image classification dataset.}
	\KwOut{The individual $individual$ with its fitness.}
	Construct a CNN with a classifier based on the information encoded in $individual$ and the given image classification dataset\; \label{alg_ini_eval_line1}
	$v_{best}\leftarrow 0$\;
	\ForEach{epoch in the given training epochs}
	{
		Train the CNN on $D_{train}$ by using the given GPU\;
		\label{alg_ini_eval_train_line1}
		$v\leftarrow$ Calculate the classification accuracy on $D_{fitness}$\;
		\label{alg_ini_eval_train_line2}
		\If{$v>v_{best}$}
		{
			$v_{best}\leftarrow v$\;
		}
	}
	Set $v_{best}$ as the fitness of $individual$\;\label{alg_ini_eval_train_line3}
	Put the identifier of $individual$ and $v_{best}$ into $Cache$\;
	\label{alg_ini_eval_train_line4}
	\textbf{Return} $individual$.

\end{algorithm}

The details of evaluating the fitness of one individual are shown in Algorithm~\ref{alg_fitness_individual_eval}. Firstly, a CNN is decoded from $individual$, and a classifier is added to this CNN (line~\ref{alg_ini_eval_line1}) based on the given image classification dataset. In the proposed algorithm, a softmax classifier~\cite{nasrabadi2007pattern} is used, and the particular number of classes is determined by the given image dataset. When decoding a CNN, a rectifier activation function~\cite{glorot2011deep} followed by a batch normalization~\cite{NIPS2017_6790} operation is added to the output of the convolutional layer, which is based on the conventions of modern CNNs~\cite{he2016deep,huang2017densely}. In addition, when the spatial number of the skip layer differs from that of the input data, a convolutional layer, which is with the unit filter and the unit stride but the special number of the feature maps, is added to the input data~\cite{he2016deep,huang2017densely}. After that, the CNN is trained by the Stochastic Gradient Descent (SGD) algorithm~\cite{bottou2012stochastic} on the training data by using the given GPU (line~\ref{alg_ini_eval_train_line1}), and the classification accuracy is calculated on the fitness evaluation data (line~\ref{alg_ini_eval_train_line2}). Note that, the use of the softmax classifier and SGD training method are based on the conventions of the deep learning community. When the training phase is finished, the best classification accuracy on the fitness evaluation data is set as the fitness of $individual$ (line~\ref{alg_ini_eval_train_line3}). Finally, the identifier and fitness of $individual$ are associated and put into $Cache$ (line~\ref{alg_ini_eval_train_line4}).

Next, the reasons for designing such an asynchronous and a cache components are given. In summary, because the training on CNNs is very time-consuming, varying from several hours to even several months depending on the particular architecture, they are designed to speed up the fitness evaluation in the proposed algorithm. Specifically, the asynchronous component is a parallel computation platform based on GPUs. Due to the computational nature of calculating the gradients, deep learning algorithms are typically placed on GPUs to speed up the training~\cite{helfenstein2012parallel}. Indeed, existing deep learning libraries, such as Tensorflow~\cite{abadi2016tensorflow} and PyTorch~\cite{paszke2017automatic}, support the calculation on multiple GPUs. However, their parallel calculations are based on the data-parallel and model-parallel pipelines. In the data-parallel pipeline, the input data is divided into several small groups, and each group is placed on one GPU for the calculation. The reason is that the limited memory of one GPU cannot effectively handle the whole data at the same time. In the model-parallel pipeline, a model is divided into several small models, and each GPU carries one small model. The reason is that the limited computational capability on one GPU cannot run a whole model. However, the designed parallel pipeline obviously does not fall into either of the pipelines, but a higher level based on them. Hence, such an asynchronous component is designed to make full use of the GPU computational resource, especially for the population-based algorithms. Furthermore, the asynchronous component is widely used in solving a large problem, if the problem can be divided into several independent sub-problems. By parallel performing these sub-problems in different computational platforms, the total processing time of the whole problem is consequently shortened. In the past, evolutionary algorithms are typically used to solve the problems of which the fitness evaluation is not time-consuming\footnote{Although there is a type of computationally expensive problems, their fitness is commonly calculated by a surrogate model to bypass the direct fitness evaluation.}, and there is no high need in developing such asynchronous components. Occasionally, they just use the built-in components based on the adopted programming languages. However, almost all such built-in components are based on CPUs, and they cannot effectively train deep neural networks, mainly because the acceleration platform for neural networks are based on GPUs. Furthermore, the fitness evaluation of each individual is independent, which just satisfies the scene of using this technique. Motivated by the reasons described above, such an asynchronous component is designed in the proposed algorithm. The cache component is also used to speed up the fitness evaluation, which is based on the following considerations: 1) the individuals surviving into the next generation do not need to evaluate the fitness again if its architecture is not changed, and 2) the architecture, which has been evaluated, could be regenerated with the mutation and crossover operations in another generation. Note that, the weight inheriting of Large-scale Evolution cannot work for the second consideration.

Commonly, a cache system should seriously treat its size and provide the details to discuss the conflicting problem resulted from the duplicate keys. In the proposed algorithm, none is necessary to be investigated. Firstly, the cache component is similar to a map data structure where each record in this component is a string combining the identifier and the fitness value of a CNN. For example, a record such as ``identifier1=98.12" denotes that the identifier is ``identifier1" and its fitness value is ``98.12". Secondly, as we have highlighted, the identifier is calculated by the 224-hash code that can generate $2^{224}$ different identifiers. In practice, the CNN architecture algorithms, including the proposed algorithm, only evaluate thousands of CNNs. Obviously, we do not need to consider the conflicting problem because the conflicting problem will not happen. Thirdly, the 224-hash code implementation used for the proposed algorithm will generate the identifier having the length of 32, and the fitness value is the classification accuracy that is denoted by a string having the length of 4. Totally, each record in the cache component is a string having the length of 37 occupying 37 bytes with the UTF-8 file encoding. Obviously, the cache file will only occupy very little disk space even though there are thousands of records. Therefore, we do not need to concern about the size of the cache component.

\subsection{Offspring Generating}
\label{sub_section_offspring_gen}
\begin{algorithm}
	\label{alg_offspring_gene}
	\caption{Offspring Generating}
	\KwIn{The population $P_t$ containing individuals with fitness, the probabity for crossover operation $p_c$, the probability for mutation operation$p_m$, the mutation operation list $l_m$, the probabilities of selecting different mutation operations $p_l$.}
	\KwOut{The offspring population $Q_t$.}
	
	$Q_t\leftarrow \emptyset$\;
	\label{alg_crossover_begin}
	\While{$|Q_t|<|P_t|$}
	{
		$p_1\leftarrow$ Randomly select two individuals from $P_t$, and from the two then select the one with the better fitness\;
		\label{alg_off_gene_p_selc}
		$p_2\leftarrow$ Repeat Line~\ref{alg_off_gene_p_selc}\;
		\label{alg_off_gene_pp_selc}
		\While{$p_2 == p_1$}
		{
			Repeat Line~\ref{alg_off_gene_pp_selc}\;
		}\label{alg_off_gene_p_selc_end}
		
		$r\leftarrow$ Randomly generate a number from $(0,1)$\;
		\label{alg_off_gene_r1}
		\eIf{$r<p_c$}
		{
			Randomly choose a point in $p_1$ and divide it into two parts\;
			\label{alg_off_cross_r1}
			Randomly choose a point in $p_2$ and divide it into two parts\;
			$o_1\leftarrow$ Join the first part of $p_1$ and the second part of $p_2$\;
			$o_2\leftarrow$ Join the first part of $p_2$ and the second part of $p_1$\;
			$Q_t\leftarrow Q_t\cup o_1 \cup o_2$\;
			\label{alg_off_cross_r2}
		}
		{
			$Q_t\leftarrow Q_t\cup p_1 \cup p_2$\;
			\label{alg_off_gene_no_cross}
		}
	}\label{alg_crossover_end}
	\ForEach{individual p in $Q_t$}
	{\label{alg_mutation_begin}
		$r\leftarrow$ Randomly generate a number from $(0,1)$\;
		\label{alg_off_mutation_r1}
		\If{$r<p_m$}
		{\label{alg_off_mutation_begin}
			$i\leftarrow$ Randomly choose a point in $p$\;
			$m\leftarrow$ Select one operation from $l_m$ based on the probabilities in $p_l$\;
			Do the mutation $m$ at the point $i$ of $p$\;
		}\label{alg_off_mutation_end}
	}\label{alg_mutation_end}
	\textbf{Return} $Q_t$.

\end{algorithm}
The details of generating the offspring are shown in Algorithm~\ref{alg_offspring_gene}, which is composed of two parts. The first is crossover (lines~\ref{alg_crossover_begin}-\ref{alg_crossover_end}) and the second is mutation (lines~\ref{alg_mutation_begin}-\ref{alg_mutation_end}). During the crossover operation, there will be totally $|P_t|$ offspring generated, where $|\cdot|$ measures the size of the collection. Specifically, two parents are selected first, and each is selected from two randomly selected individuals based on the better fitness (lines~\ref{alg_off_gene_p_selc}-\ref{alg_off_gene_p_selc_end}). This selection is known as the binary tournament selection~\cite{miller1995genetic}, which is popular in GAs for single-objective optimization. Once the parents are selected, a random number is generated~(line~\ref{alg_off_gene_r1}), determining whether the crossover will be done or not. If the generated number is not below the predefined crossover probability, these two parent individuals are put into $Q_t$ as the offspring (line~\ref{alg_off_gene_no_cross}). Otherwise, each parent individual is randomly split into two parts, and the two parts from the two parent individuals are swapped to formulate two offspring (lines~\ref{alg_off_cross_r1}-\ref{alg_off_cross_r2}). During the mutation operation, a random number is generated first (line~\ref{alg_off_mutation_r1}), and the mutation operation is performed on the current individual if the generated number is below than $p_m$ (lines~\ref{alg_off_mutation_begin}-\ref{alg_off_mutation_end}). When mutating an individual, a position (denoted as $i$) is randomly selected from the current individual, and one particular mutation operation (denoted as $m$) is selected from the provided mutation list based on the probabilities defined in $p_l$. Then, $m$ is performed on the position $i$. In the proposed algorithms, the available mutation operations defined in the mutation list are:
\begin{itemize}
	\item Adding a skip layer with random settings;
	\item Adding a pooling layer with random settings;
	\item Remove the layer at the selected position; and
	\item Randomly changing the parameter values of the building block at the selected position.
\end{itemize}

Next, the motivation of designing such a crossover operator and selecting a mutation operation based on a provided probability list is given. Firstly, the designed crossover operator is inspired by the one-point crossover~\cite{srinivas1994genetic} in traditional GAs. However, the one-point crossover was designed only for the individuals with the equal lengths. The designed crossover operator is used for the individuals with unequal lengths. Although the designed crossover operator is simple, it dose improve the performance in discovering the architectures of CNNs, which will be experimentally proved in Section~\ref{section_exp_results}. Secondly, existing algorithms use an equal probability for choosing the particular mutation operation. In the proposed algorithm, the provided mutation operations are selected with different probabilities. Specifically, we provide a higher probability for the ``adding a skip layer" mutation, which will have a higher probability to increase the depths of CNNs. For other mutation operations, we still employ the equal probabilities. The motivation behind this design is that a deeper CNN will have a more powerful capability, as mentioned previously. Although the ``adding a pooling layer'' is also capable of increasing the depth of CNNs, the dimension of input data will be reduced to half by using one pooling layer, which results in the unavailability of the discovered CNN. To this end, we do not put a higher probability on it.

Please note that the optimal depth of the CNN is achieved by the proposed mutation operators. As shown above, there are four different types of mutation operators designed in the proposed algorithm. The optimal depth can be found through the first three mutation operators. Particularly, the first two operators provide a chance to increase the depth of the CNN (denoted both as the ``depth-increasing operator"), while the third mutation operator provides a chance to decrease the depth (denoted it as the ``depth-decreasing operator"). For example, if the optimal depth of the CNN is $N$, after the random initialization, some individuals may have the length (denoting the depth of the CNNs) smaller than $N$, while others are with length greater than $N$. During evolution, each individual has a chance to be mutated. If the individual whose length is greater than $N$ and the depth-decreasing operator is selected, the individuals' length will be decreased towards $N$, and vice versa. In summary, the optimal depth is obtained by performing the mutation operations on the individuals randomly initialized with different lengths.

\subsection{Environmental Selection}
\label{sub_section_environmental_selection}
\begin{algorithm}
	\label{alg_environ_sele}
	\caption{Environmental Selection}
	\KwIn{The parent population $P_t$, the offspring population $Q_t$}
	\KwOut{The population for the next generation $P_{t+1}$.}
	$P_{t+1}\leftarrow\emptyset$\;
	\While{$|P_{t+1}|<|P_t|$}
	{\label{alg_env_sele_l1}
		$p_1, p_2\leftarrow$ Randomly select two individuals from $Q_t\cup P_t$\;
		$p\leftarrow$ Select the one who has a better fitness from $\{p_1,p_2\}$\;
		$P_{t+1}\leftarrow P_{t+1}\cup p$\;
	}\label{alg_env_sele_l2}
	$p_{best}\leftarrow$ Find the individual with the best fitness from $Q_t\cup P_t$\;
	\label{alg_env_sele_l3}
	\If{$p_{best}$ is not in $P_{t+1}$}
	{
		Replace the one who has the worst fitness in $P_{t+1}$ by $p_{best}$; 
	}\label{alg_env_sele_l4}
	\textbf{Return} $P_{t+1}$.

\end{algorithm}
Algorithm~\ref{alg_environ_sele} shows the details of the environmental selection. Firstly, $|P_{t}|$ individuals are selected from the current population ($Q_t\cup P_t$) by using the binary tournament selection, and then these selected individuals are placed into the next population (denoted $P_{t+1}$) (lines~\ref{alg_env_sele_l1}-\ref{alg_env_sele_l2}). Secondly, the best individual is selected and to check whether it has been placed into $P_{t+1}$. If not, it will replace the worst individual in $P_{t+1}$ (lines~\ref{alg_env_sele_l3}-\ref{alg_env_sele_l4}).

In principle, only selecting the top $|P_t|$ best individuals for the next generation easily causes the premature phenomenon~\cite{michalewicz1996genetic}, which will lead the algorithm trap into a local optimum~\cite{goldberg1988genetic,davis1991handbook}. If we do not explicitly select the best individuals for the next generation, the algorithm will not convergence. In principle, a desirable population should contain not only good but also the relatively bad ones for enhancing the diversity~\cite{anderson2005practical,malik2014preventing}. To this end, the binary tournament selection is typically used for such a purpose~\cite{zhang2003novel,miller1995genetic}. However, only using the binary tournament selection may miss the best individual, resulting in the algorithm not towards a better direction of the evolution. Hence, we explicitly add the best individual to the next population, which is a particular elitism strategy in evolutionary algorithms~\cite{bhandari1996genetic}. 

Note that the binary tournament selection, used in the proposed algorithm (they are used in the offspring generating and environmental selection phases), is used with replacement. Based on our previous work~\cite{xie2009sampling}, the tournament selection with or without replacement almost has no different bias to the final performance, and the replacement mechanism used in the proposed algorithm is just to follow the common practice~\cite{deb2002fast,sun2018igd,transferjiang2018}.

\section{Experiment Design}
\label{section_exp_settings}

In order to evaluate the performance of the proposed algorithm, a series of experiments has been conducted on image classification tasks. Specifically, the peer competitors chosen to compare with the proposed algorithm are introduced in Subsection~\ref{sub_section_peer_competitor}. Then, the used benchmark datasets are detailed in Subsection~\ref{sub_section_benchmark}. Finally, the parameter settings of the proposed algorithm are shown in Subsection~\ref{sub_section_parameter}.

\subsection{Peer Competitors}
\label{sub_section_peer_competitor}
In order to show the effectiveness and efficiency of the proposed algorithm, the state-of-the-art algorithms are selected as the peer competitors to the proposed algorithm. Particularly, the peer competitors are chosen from three different categories. 

The first refers to the state-of-the-art CNNs that are manually designed, including ResNet~\cite{he2016deep}, DenseNet~\cite{huang2017densely}, VGGNet~\cite{simonyan2014very}, Maxout~\cite{goodfellow2013maxout}, Network in Network~\cite{lin2013network}, Highway Network~\cite{srivastava2015highway} and All-CNN~\cite{springenberg2014striving}. Specifically, two versions of ResNet, i.e., the ResNet models with a depth of $110$ and $1,202$, are used for the comparison. For the convenience, they are named ResNet (depth=$110$) and ResNet (depth=$1,202$), respectively. Among all the ResNet versions investigated in the seminal paper~\cite{he2016deep}, ResNet (depth=$110$) is the winner in terms of the classification accuracy on CIFAR10, and ResNet (depth=$1,202$) is the most complex version, which is the main reason for selecting them for the comparison. For the DenseNet, we use the version named ``DenseNet-BC'' that achieved the promising classification accuracy but also has the least number of parameters among its all variants~\cite{huang2017densely}. Note that, most algorithms from this category are the champions in the large-scale visual recognition challenges in recent years~\cite{ILSVRC15}.

The second and the third contain the CNN architecture design algorithms from the ``automatic+manually tuning'' and the ``automatic'' categories, respectively. Specifically, Genetic CNN~\cite{xie2017genetic}, Hierarchical Evolution~\cite{liu2017hierarchical}, EAS~\cite{zoph2016neural}, Block-QNN-S~\cite{zhong2017practical}, DARTS~\cite{liu2018darts}, and NSANet~\cite{zoph2018learning} belong to the second category, while Large-scale Evolution~\cite{real2017large}, CGP-CNN~\cite{suganuma2017genetic}, NAS~\cite{zoph2016neural} and MetaQNN~\cite{baker2016designing} fall into the third category. In the comparison, we choose two versions of NASNet, i.e., NASNet-B and NASNet-A+cutout, which are the ones having the least number of parameters and the best classification accuracy, respectively. Specifically, the ``cutout'' refers to a regularization method~\cite{devries2017improved} used in the training of CNNs, which could improve the final performance.

Please note that, the proposed algorithm mainly focuses on providing an ``automatic'' way to design promising CNN architectures for users who have no rich domain knowledge of tuning CNN architectures. Based on the ``No Free Lunch Theorems~\cite{wolpert1997no}'', we should realize that CNNs whose architectures are designed with manual tuning should have better classification accuracy than the ``automatic'' ones including the proposed algorithm. Indeed, only the comparison to the CNN architecture designs from the ``automatic'' category is fair to the proposed algorithm. In this experiment, we still would like to provide the extensive comparisons to the CNN architectures designed with domain knowledge,  to show the efficiency and effectiveness of the proposed algorithm among all existing state-of-the-art CNNs.

\subsection{Benchmark Datasets}
\label{sub_section_benchmark}
The CIFAR10 and CIFAR100 benchmark datasets~\cite{krizhevsky2009learning} are chosen as the image classification tasks in the experiments. The reasons for choosing them are that 1) both datasets are challenging in terms of the image sizes, categories of classification and the noise as well as rotations in each image; and 2) they are widely used to measure the performance of deep learning algorithms, and most of the chosen compared algorithms have publicly reported their classification accuracy on them. 

\begin{figure}[!htp]
	\centering
	\subfloat[CIFAR10]{\includegraphics[width=0.95\columnwidth]{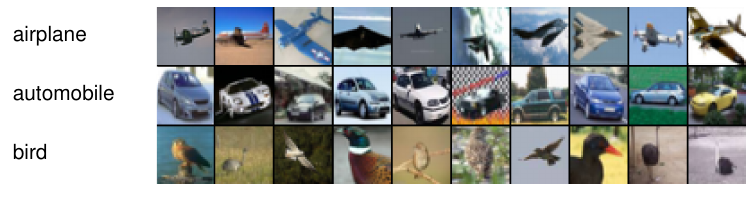}\label{fig_cifar10_example}}
	\hfil
	\subfloat[CIFAR100]{\includegraphics[width=0.95\columnwidth]{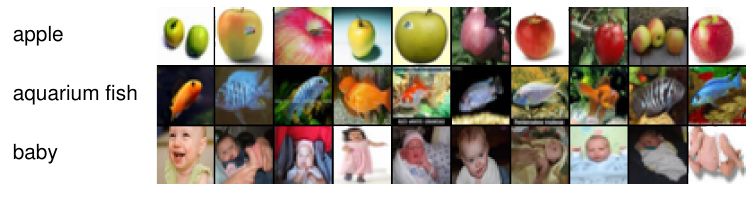}\label{fig_cifar100_example}}
	\caption{Examples from CIFAR10 (shown in Fig.~\ref{fig_cifar10_example}) and CIFAR100 (shown in Fig.~\ref{fig_cifar100_example}). Each row denotes the images from the same category, and the first column displays the corresponding category label.}\label{fig_benchmark_example}
\end{figure}

Specifically, the CIFAR10 dataset is an image classification benchmark for recognizing $10$ classes of natural objects, such as air planes and birds. It consists of $60,000$ RGB images in the dimension of $32\times 32$. In addition, there are $50,000$ images and $10,000$ images in the training set and the testing set, respectively. Each category has an equal number of images. On the other hand, the CIFAR100 dataset is similar to CIFAR10, except it has $100$ classes. In order to have a quick glance on both datasets, we randomly select three classes from each benchmark dataset, and then randomly select $10$ images from each class. These selected images are shown in Fig.~\ref{fig_benchmark_example}. Specifically, Fig.~\ref{fig_cifar10_example} shows the images from CIFAR10, while Fig.~\ref{fig_cifar100_example} displays those from CIFAR100. The left column in Fig.~\ref{fig_benchmark_example} shows the class names of the images in the same rows. It can be observed that, the objects to be classified in these benchmark datasets typically occupy different areas of the whole image, and their positions are not the same in different images. Their variations are also challenging for the classification algorithms.   

In the experiments, the training set is split into two parts. The first part accounts for $90\%$ to serve as the training set for training the individuals, while the remaining images serve as the fitness evaluation set for evaluating the fitness. In addition, the images are augmented during the training phases. In order to do a fair comparison, we employ the same augmentation routine as those often used in peer competitors, i.e., each direction of one image is padded by four zeros pixels, and then an image with the dimension of $32\times32$ is randomly cropped. Finally, a horizontal flip is randomly performed on the cropped image with a probability of 0.5~\cite{he2016deep,huang2017densely}. 

To date, most architecture discovering algorithms do not perform experiments on the CIFAR100 dataset due to its large number of classes. In order to show the superiority of our proposed algorithm over the peer competitors, we perform experiments on CIFAR100 and report the results.

\subsection{Parameter Settings}
\label{sub_section_parameter}
As have been discussed, the main objective of this paper is to design an automatic architecture discovering algorithm for researchers without domain expertise in CNNs. To further improve the applicability of the proposed algorithm, we design it in a way that potential users are not required to have expertise in evolutionary algorithms either. Hence, we simply set the parameters of the proposed algorithm based on the conventions. Specifically, the probabilities of crossover and mutation are set to $0.9$ and $0.2$, respectively, as suggested in~\cite{back1996evolutionary}. During the training of each individual, the routine in~\cite{he2016deep} is employed, i.e., the SGD with the learning rate of $0.1$ and the momentum of $0.9$ are used to train $350$ epochs and the learning rate is decayed by a factor of $0.1$ at the $1$-st, $149$-th and $249$-th epochs. Indeed, most peer competitors are also based on this training routine. When the proposed algorithm terminates, we choose the one with the best fitness value, and then train it up to $350$ epochs on the original training set. At last, the classification accuracy on the testing set is tabulated to compare with peer competitors. In addition, the available numbers of feature maps are set to $\{64, 128, 256\}$ based on the settings employed by the state-of-the-art CNNs. As for the probabilities of the four mutation operations shown in Subsection~\ref{sub_section_offspring_gen}, we set the normalized probability of increasing the depth to $0.7$, while others share the equal probabilities. Theoretically, any probability can be set for the adding mutation by keeping it higher than that of others. In addition, the population size and the number of generations are all set to $20$, and the similar settings are also employed by the peer competitors. Ideally, a larger population size and a larger maximal generation number should result in a better performance, but expectedly also consume more computational resources. However, such optimal settings are not within the scope of this paper since the current settings, as commonly adopted by other papers, have easily outperformed most of the peer competitors based on the results shown in Table~\ref{tab_overview}. Noting that, the experiments of the proposed algorithm are performed on three GPU cards with the same model of Nvidia GeForce GTX 1080 Ti.

\begin{table*}
	\renewcommand{\arraystretch}{0.8}
	\caption{The comparisons between the proposed algorithm and the state-of-the-art peer competitors in terms of the classification accuracy (\%), number of parameters and the taken GPU days on the CIFAR10 and CIFAR100 benchmark datasets.}
	\label{tab_overview}
	\center
	\begin{tabular}{c|l|c|c|c|c|r}
		\hline
		& &\textbf{CIFAR10} & \textbf{CIFAR100} & \textbf{\# Parameters} & \textbf{GPU days} & \textbf{Manual assistance?} \\
		\hline
		\multirow{8}{*}{\shortstack[l]{\textbf{manually designed}}}& ResNet (depth=110) & 93.57 & 74.84 & 1.7M & -- & completely needed\\
		& ResNet (depth=1,202) & 92.07 & 72.18 & 10.2M & -- & completely needed\\
		& DenseNet-BC & 95.49 & 77.72 & 0.8M & -- & completely needed\\
		& VGGNet & 93.34 & 71.95 & 20.04M & -- & completely needed\\
		& Maxout & 90.70 & 61.40 & -- & --& completely needed\\
		& Network in Network & 91.19 & 64.32 & -- & -- & completely needed\\
		& Highway Network & 92.40 & 67.66 & -- & -- & completely needed\\
		& All-CNN & 92.75 & 66.29 & 1.3M & -- & completely needed\\
		\hline
		\multirow{6}{*}{\shortstack[l]{\textbf{automatic + manually tuning}}} & Genetic CNN & 92.90 &70.97 & -- & 17 & partially needed \\
		& Hierarchical Evolution & 96.37 & -- & -- & 300 & partially needed \\
		& EAS & 95.77 & -- & 23.4M & 10 & partially needed \\
		& Block-QNN-S & 95.62 & 79.35 & 6.1M & 90 & partially needed \\
		& DARTS & 97.00 & -- & 3.3M & 4 & partially needed \\
		& NASNet-B & 96.27 & -- & 2.6M & 2,000 & partially needed\\
		& NASNet-A + cutout & 97.60 & -- & 27.6M & 2,000 & partially needed\\
		\hline
		\multirow{8}{*}{\shortstack[l]{\textbf{automatic}}} & Large-scale Evolution & 94.60 & -- & 5.4M & 2,750 & completely not needed \\
		& Large-scale Evolution & -- & 77.00 & 40.4M & 2,750 & completely not needed \\
		& CGP-CNN & 94.02 & -- & 1.68M & 27 & completely not needed \\
		& NAS & 93.99 & -- & 2.5 M & 22,400 & completely not needed \\
		& Meta-QNN & 93.08 & 72.86 & -- & 100 & completely not needed \\
		\cline{2-7}
		& CNN-GA & 95.22 & -- & 2.9M & 35 & completely not needed \\
		& CNN-GA & -- & 77.97 & 4.1M  & 40 & completely not needed\\
		& CNN-GA + cutout& 96.78 & -- & 2.9M & 35 & completely not needed \\
		& CNN-GA  + cutout& -- & 79.47 & 4.1M  & 40 & completely not needed\\
		\hline
	\end{tabular}
\end{table*}

\section{Experimental Results and Analysis}
\label{section_exp_results}

In this section, we firstly show an overview of the comparison results between the proposed algorithm and the chosen peer competitors. Then, the evolutionary trajectories of the proposed algorithm are shown, which helps readers to better understand the proposed algorithm during the process of discovering the best CNN architecture and the appropriateness of setting to the generation number.

\subsection{Overall Results}
Because the state-of-the-art CNNs in the first category of peer competitors are manually-designed, we mainly compare the classification accuracy and the number of parameters. For the algorithms in the other two categories, we compare the ``GPU days'' used to discover the corresponding CNN, in addition to the classification accuracy and the number of parameters. Particularly, the unit GPU day means the algorithm has performed one day on one GPU when the algorithm terminates, which reflects the computational resource consumed by these algorithms. For the convenience of summarizing the comparison results, we use the name of the architecture discovering algorithm as the name of the discovered CNN when comparing the classification accuracy and the number of parameters between peer competitors. For example, the proposed algorithm is named CNN-GA which is an architecture discovering algorithm, so its discovered CNN is titled as CNN-GA when comparing it to the chosen state-of-the-art CNNs.

The comparison results between the proposed algorithm and the peer competitors are shown in Table~\ref{tab_overview}. In Table~\ref{tab_overview}, the peer competitors are grouped into three different blocks based on the categories, and the first column shows the category names. As a supplement, the last column also provides the information regarding how much manual assistance the corresponding CNN requires during discovering the architectures of CNNs. Additionally, the second column denotes the names of the peer competitors. The third and fourth columns refer to the classification accuracies on the CIFAR10 and CIFAR100 datasets, while the fifth column displays the numbers of parameters in the corresponding CNNs. The sixth column shows the used GPU days which are only applicable to the semi-automatic and automatic algorithms. The symbol ``--'' implies there is no result publicly reported by the corresponding algorithm. For ResNet (depth=110), ResNet (depth=1,202), DenseNet, VGG, All-CNN, EAS and Block-QNN-S, they have the same number of parameters on both the CIFAR10 and CIFAR100 datasets, which is caused by the similar CNNs achieving the best classification accuracy on both datasets. Note that, the results of the peer competitors shown in Table~\ref{tab_overview} are all from their respective seminal papers\footnote{It is exceptional for VGG because it does not perform experiments on CIFAR10 and CIFAR100 in its seminal paper. Its results displayed in Table~\ref{tab_overview} is derived from~\cite{suganuma2017genetic}.}. For the proposed algorithm, the best CNN discovered by CNN-GA is selected from the population in the last generation, and then trained independently for five times. The best classification accuracy is selected from the five results to show in Table~\ref{tab_overview}, which follows the conventions of its peer competitors~\cite{xie2017genetic,liu2017hierarchical,zoph2016neural,zhong2017practical,real2017large,suganuma2017genetic,baker2016designing}. Inspired by the performance improvement of NASNet given by the cutout regularization, we also report the results by performing CNN-GA with the cutout, which is denoted as ``CNN-GA+cutout'' in Table~\ref{tab_overview}. Specifically, we firstly choose the best CNN-GA architectures found in CIFAR10 and CIFAR100, respectively, then retrained these architectures with the cutout on the training set, and then report the classification accuracy on the testing set.

In this experiment, we try our best to do a fair comparison by using the same training process as well as the same data augmentation method as those of the chosen peer competitors. However, because of multiple peer competitors aiming at finding the best classification accuracy, different training process and data augmentation methods are adopted by them. For example, NASNet employed a revised path drop technique to improve the classification accuracy. Because the revised path drop technique is not publicly available, we only keep the proposed algorithm with the same training process and data augmentation methods as those of the most peer competitors. Therefore, we should keep in mind that solely comparing the classification accuracy is not fair to the proposed algorithm either.

For the peer competitors in the first category, CNN-GA obtains $3.15\%$ and $1.88\%$ improvements in terms of the classification accuracy on the CIFAR10 dataset over ResNet (depth=1,202) and VGG, respectively, while using merely $28\%$ and $14\%$ of their respective parameters. The number of parameters in CNN-GA is more than that of ResNet (depth=110) and All-CNN on the CIFAR10 dataset, but CNN-GA shows the best classification accuracy among them. For DenseNet-BC, which is the most recent state-of-the-art CNN, CNN-GA shows slightly worse classification accuracy on CIFAR10, but on the CIFAR100 that is a more complex benchmark dataset than CIFAR10, CNN-GA shows slightly better classification accuracy. In addition, CNN-GA achieves the highest classification accuracy among Maxout, Network in Network and Highway Network on the CIFAR10 dataset. On the CIFAR100 dataset, CNN-GA employs $52\%$, $85\%$ and $40\%$ fewer parameters compared to ResNet (depth=1202), DenseNet and VGG, respectively, but even achieves $5.79\%$, $1.39\%$ and $6.02\%$ improvements over the respective classification accuracy. CNN-GA also outperforms ResNet (depth=110), Maxout, Network in Network, Highway Network and All-CNN in terms of the classification accuracy, although it uses a larger network than that of RestNet (depth=101) and All-CNN. In summary, CNN-GA achieves the best classification accuracy on both the CIFAR10 and CIFAR100 datasets among the state-of-the-art CNNs manually designed. Additionally, it also employs a much fewer number of parameters than most of the state-of-the-art CNNs in the first category.

For the peer competitors in the second category, the classification accuracy of CNN-GA is better than that of Genetic CNN. CNN-GA shows $1.15\%$ lower classification accuracy than that of Hierarchical Evolution; however, CNN-GA takes only one-tenth of the GPU days compared with that of Hierarchical Evolution. Compared to Block-QNN-S, CNN-GA shows slightly worse classification accuracy on the CIFAR10 and CIFAR100 datasets, while CNN-GA consumes about half of the number of the parameters and also half of the GPU days compared to those of Block-QNN-S. Furthermore, the classification accuracy of CNN-GA is competitive to that of EAS, and CNN-GA employs a significantly smaller number of parameters than that of EAS ($87\%$ fewer parameters used by CNN-GA than that of EAS). In addition, NASNet-B shows slightly better classification accuracy on CIFAR10 than CNN-GA (96.20\% v.s. 95.22\%), while CNN-GA only consumes about 1/55 GPU days of that consumed by NASNet-B (35 GPU days v.s., 2,000 GPU days). Compared to NASNet-B, NASNet-A+cutout improves a classification accuracy of 1.33\% with a CNN having 27.6M parameters, by consuming 2,000 GPU days; but the CNN-GA+cutout obtains a 1.76\% classification accuracy improvement, while still with the same CNN having 2.9M parameters and consumes 35 GPU days. Although CNN-GA targets at designing promising CNN architectures without any CNN domain knowledge, it still shows comparable classification accuracy among the peer competitors in this ``automatic+manually tuning'' category. Compared to CNN-GA, the major limitation of the algorithms in this category is the requirement of extended expertise when they are used to solve real-world tasks. For example, EAS requires a manually-tuned CNN on the given dataset, and then EAS is used to refine the tuned CNN. If the tuned CNN is not with a promising performance, the developed EAS would perform not well either at the end. In addition, the CNNs discovered by Hierarchical Evolution and Block-QNN-S cannot be directly used. They must be inserted into a larger CNN manually-designed in advance. If the larger network is not designed properly, the final performance of Hierarchical Evolution and Block-QNN-S would also perform poorly. In addition to the CNN framework that is required by all versions of NASNet, NASNet also employed a revised path-dropping technique to improve its performance. In addition, as can be seen in Table~\ref{tab_overview}, DARTS achieves the similar performance as well as the number of parameters as CNN-GA on CIFAR10, while consumes much fewer GPU days than CNN-GA dose. However, DARTS requires a manually-designed large CNN as a building block, and then using the differential method, i.e., the SGD, to choose the best path among the large CNNs. If the provided CNN is smaller than the optimal one, DARTS can never find the optimal CNN architecture. The proposed CNN-GA bears no such limitations, although it consumes more GPU days than DARTS dose.

For the peer competitors in the third category, CNN-GA performs better than Large-scale Evolution and CGP-CNN, and much better than NAS and Meta-QNN on the CIFAR10 dataset in terms of the classification accuracy. Meanwhile, CNN-GA also shows superiority in the classification accuracy over Large-scale Evolution and Meta-QNN on the CIFAR100 dataset. Furthermore, CNN-GA only has $2.9$M parameters on the CIFAR10 dataset, which is almost half of those of Large-scale Evolution. On the CIFAR100 dataset, CNN-GA has $4.1$M parameters, which saves $90\%$ parameters compared to those of Large-scale evolution which requires $40.4$M parameters~\cite{real2017large}. Furthermore, CNN-GA takes only $35$ GPU days on the CIFAR10 dataset and $40$ GPU days on the CIFAR100 dataset, while Large-scale Evolution consumes $2,750$ GPU days on the CIFAR10 dataset and another $2,750$ GPU days on the CIFAR100 dataset. Even more, NAS employs $22,400$ GPU days on the CIFAR10 dataset. Based on the number of parameters and GPU days taken, it can be summarized that CNN-GA shows a promising performance by using $90\%$ simpler architectures on $99\%$ less computational resource than the average numbers of competitors in this category\footnote{The number is calculated by summing up the corresponding number of each peer competitor in this category, and then normalized by the numbers of available classification results. For example, the total GPU days of the peer competitors are $28,027$, and there are only six available classification results. Hence, the average GPU days are $4,671$. In the same way, the average GPU days took by CNN-GA are found to be $37.5$.}. 

In summary, CNN-GA outperforms most of the state-of-art CNNs manually designed and all the automatic architecture discovering algorithms in terms of not only the classification accuracy, but also the number of parameters and the employed computational resource. Although the state-of-the-art ``automatic+manually-tuning'' architecture discovering algorithms show similar (to slightly better) classification accuracy to that of CNN-GA, CNN-GA is automatic and dose not require users to have any expertise in CNNs when solving real-world tasks, which is the main purpose of our work in this paper.

To make the readers intuitively understand the efficacy of the proposed one on the image classification tasks, the representative of the conventional image classification methods~\cite{dalal2005histograms}, i.e., the Histograms of Oriented Gradient (HOG) descriptor with the Support Vector Machine (SVM) (in short, denoted as HOG+SVM), is used to compare the classification accuracy on CIFAR10 and CIFAR100. Particularly, we use the source code of HOG+SVM downloaded from the Github\footnote{https://github.com/subicWang/HOG-SVM-classifer}, and the resulted classification accuracies of CIFAR10 and CIFAR100 are 51.22\% and 43.90\%, respectively. As compared with the results of the proposed algorithm, which are 96.78\% and 79.47\%, respectively, it is easy to conclude that the proposed algorithm can outperform the HOG+SVM with the significant improvement.

\subsection{Evolutionary Trajectories}
\begin{figure}[!htp]
	\centering
	\includegraphics[width=\columnwidth]{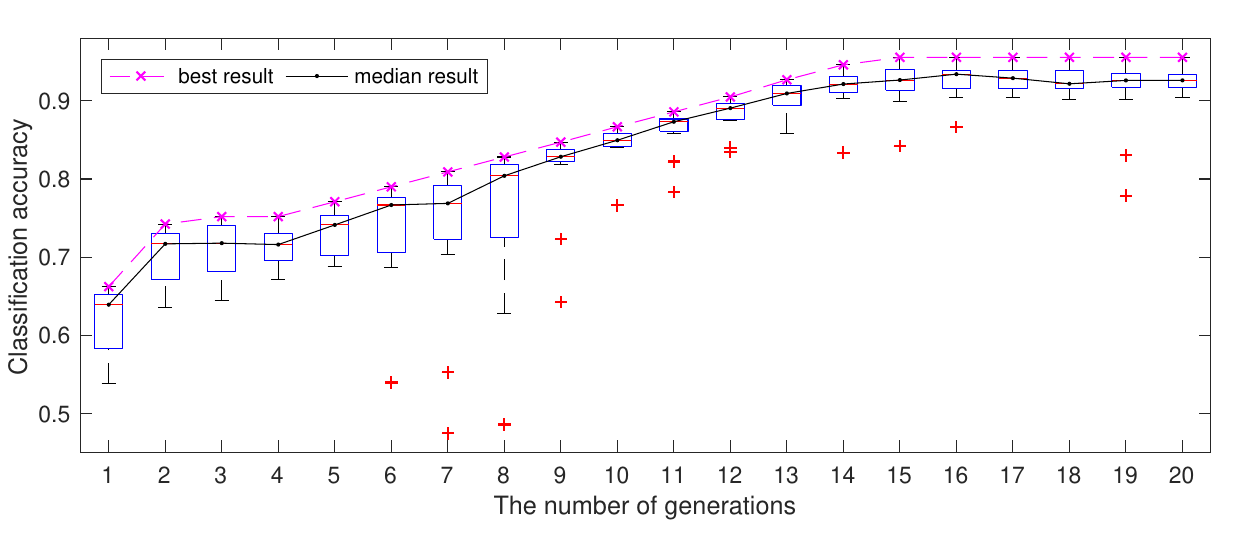}\\
	\caption{The evolutionary trajectory of the proposed algorithm in discovering the best architecture of CNN on the CIFAR10 dataset.}\label{fig_trajectory}
\end{figure}
In order to better understand the details of the proposed algorithm in discovering the architectures of CNNs and the appropriateness of setting 20 as the generation number, the evolutionary trajectory of the proposed algorithm on CIFAR10 dataset is shown in Fig.~\ref{fig_trajectory}. To achieve this, we firstly collect the individuals selected by the environmental selection in each generation, and then use the boxplot~\cite{williamson1989box} to show the statistics in terms of the classification accuracy. Meanwhile, we also connect the best and median classification accuracy during each generation by using a dashed line and a solid line, respectively. In Fig.~\ref{fig_trajectory}, the horizon axis denotes the number of generations, while the vertical axis denotes the classification accuracy.

As shown in Fig.~\ref{fig_trajectory}, both the best and median classification accuracy increase as the evolution progresses. By investigating the height of each box, it can also be observed that the variation of the classification accuracy during each generation becomes smaller and smaller, which implies the evolution towards a steady state in discovering the architectures of CNNs on CIFAR10 dataset. In addition, the classification accuracy increases sharply from the first generation to the second generation, which is due to the random initialization of the population at the beginning of the evolution. From the second to the fourth generations, the improvement of the classification accuracy becomes much smaller than the previous generations, and it sharply increases immediately after until the $15$-th generation. Since then, the classification accuracy does not change too much until the evolution terminates, which also implies that the setting of $20$ generations in this particular case is reasonable because the proposed algorithm well converges with this setting.

\section{Conclusions and Future Work}
\label{section_conclusion}
The objective of this paper is to propose an automatic architecture design algorithm for CNNs by using the GA (in short named CNN-GA), which is capable of discovering the best CNN architecture in addressing image classification problems for the users who have no expertise in tuning CNN architectures. This goal has been successfully achieved by designing a new encoding strategy for the GA to encode arbitrary depths of CNNs, incorporating the skip connections to promote deeper CNNs to be produced during the evolution and developing a parallel as well as a cache component to significantly accelerate the fitness evaluation given a limited computational resource. The proposed algorithm is examined on two challenging benchmark datasets, and compared with 18 state-of-the-art peer competitors, including eight manually-designed CNNs, six automatic+manually tuning and four automatic algorithms discovering the architectures of CNNs. The experimental results show that CNN-GA outperforms almost all the manually-designed CNNs as well as the automatic peer competitors, and shows competitive performance with respect to automatic+manually tuning peer competitors in terms of the best classification accuracy. The CNN discovered by CNN-GA has a much smaller number of parameters than those of most peer competitors. Furthermore, CNN-GA also employs significantly less computational resource than most automatic and automatic+manually tuning peer competitors. CNN-GA is also completely automatic, and users can directly use it to address their own image classification problems whether or not they have domain expertise in CNNs or GAs. Moreover, the CNN architecture designed by CNN-GA on CIFAR10 also shows the promising performance when it is transferred to the ImageNet dataset.

In CNN-GA, two components have been designed to speed up the fitness evaluation, and much computational resource has been saved. However, the computational resource employed is still fairly large than those of GAs in solving traditional problems. In the field of solving expensive optimization problems, several algorithms based on evolutionary computation techniques have been developed. In future, we will place efforts on developing effective evolutionary computation methods to significantly speed up the fitness evaluation of CNNs.

\ifCLASSOPTIONcaptionsoff
  \newpage
\fi



\begin{thebibliography}{10}
	\providecommand{\url}[1]{#1}
	\csname url@samestyle\endcsname
	\providecommand{\newblock}{\relax}
	\providecommand{\bibinfo}[2]{#2}
	\providecommand{\BIBentrySTDinterwordspacing}{\spaceskip=0pt\relax}
	\providecommand{\BIBentryALTinterwordstretchfactor}{4}
	\providecommand{\BIBentryALTinterwordspacing}{\spaceskip=\fontdimen2\font plus
		\BIBentryALTinterwordstretchfactor\fontdimen3\font minus
		\fontdimen4\font\relax}
	\providecommand{\BIBforeignlanguage}[2]{{%
			\expandafter\ifx\csname l@#1\endcsname\relax
			\typeout{** WARNING: IEEEtran.bst: No hyphenation pattern has been}%
			\typeout{** loaded for the language `#1'. Using the pattern for}%
			\typeout{** the default language instead.}%
			\else
			\language=\csname l@#1\endcsname
			\fi
			#2}}
	\providecommand{\BIBdecl}{\relax}
	\BIBdecl
	
	\bibitem{lecun2015deep}
	Y.~LeCun, Y.~Bengio, and G.~Hinton, ``Deep learning,'' \emph{Nature}, vol. 521,
	no. 7553, pp. 436--444, 2015.
	
	\bibitem{krizhevsky2012imagenet}
	A.~Krizhevsky, I.~Sutskever, and G.~E. Hinton, ``Imagenet classification with
	deep convolutional neural networks,'' in \emph{Advances in Neural Information
		Processing Systems}, 2012, pp. 1097--1105.
	
	\bibitem{sainath2013deep}
	T.~N. Sainath, A.-r. Mohamed, B.~Kingsbury, and B.~Ramabhadran, ``Deep
	convolutional neural networks for lvcsr,'' in \emph{2013 IEEE International
		Conference on Acoustics, Speech and Signal Processing}.\hskip 1em plus 0.5em
	minus 0.4em\relax IEEE, 2013, pp. 8614--8618.
	
	\bibitem{sutskever2014sequence}
	I.~Sutskever, O.~Vinyals, and Q.~V. Le, ``Sequence to sequence learning with
	neural networks,'' in \emph{Advances in Neural Information Processing
		Systems}, 2014, pp. 3104--3112.
	
	\bibitem{alphago}
	C.~Clark and A.~Storkey, ``Training deep convolutional neural networks to play
	go,'' in \emph{Proceedings of the 32nd International Conference on Machine
		Learning}, Lille, France, 2015, pp. 1766--1774.
	
	\bibitem{simonyan2014very}
	K.~Simonyan and A.~Zisserman, ``Very deep convolutional networks for
	large-scale image recognition,'' in \emph{Proceedings of the 32nd
		International Conference on Machine Learning}, Lille, France, 2015.
	
	\bibitem{szegedy2015going}
	C.~Szegedy, W.~Liu, Y.~Jia, P.~Sermanet, S.~Reed, D.~Anguelov, D.~Erhan,
	V.~Vanhoucke, A.~Rabinovich \emph{et~al.}, ``Going deeper with
	convolutions,'' in \emph{Proceedings of 2015 IEEE Conference on Computer
		Vision and Pattern Recognition}, Boston, MA, USA, 2015, pp. 1--9.
	
	\bibitem{he2016deep}
	K.~He, X.~Zhang, S.~Ren, and J.~Sun, ``Deep residual learning for image
	recognition,'' in \emph{Proceedings of 2016 IEEE Conference on Computer
		Vision and Pattern Recognition}, Las Vegas, NV, USA, 2016, pp. 770--778.
	
	\bibitem{huang2017densely}
	G.~Huang, Z.~Liu, K.~Q. Weinberger, and L.~van~der Maaten, ``Densely connected
	convolutional networks,'' in \emph{Proceedings of 2017 IEEE Conference on
		Computer Vision and Pattern Recognition}, Honolulu, HI, USA, 2017, pp.
	2261--2269.
	
	\bibitem{xie2017genetic}
	L.~Xie and A.~Yuille, ``Genetic {CNN},'' in \emph{Proceedings of 2017 IEEE
		International Conference on Computer Vision}, Venice, Italy, 2017, pp.
	1388--1397.
	
	\bibitem{liu2017hierarchical}
	H.~Liu, K.~Simonyan, O.~Vinyals, C.~Fernando, and K.~Kavukcuoglu,
	``Hierarchical representations for efficient architecture search,'' in
	\emph{Proceedings of 2018 Machine Learning Research}, Stockholm, Sweden,
	2018.
	
	\bibitem{cai2018efficient}
	H.~Cai, T.~Chen, W.~Zhang, Y.~Yu, and J.~Wang, ``Efficient architecture search
	by network transformation,'' in \emph{Proceedings of the 2018 AAAI Conference
		on Artificial Intelligence}, Louisiana, USA, 2018.
	
	\bibitem{zhong2017practical}
	Z.~Zhong, J.~Yan, and C.-L. Liu, ``Practical network blocks design with
	q-learning,'' in \emph{Proceedings of the 2018 AAAI Conference on Artificial
		Intelligence}, Louisiana, USA, 2018.
	
	\bibitem{zoph2018learning}
	B.~Zoph, V.~Vasudevan, J.~Shlens, and Q.~V. Le, ``Learning transferable
	architectures for scalable image recognition,'' in \emph{Proceedings of the
		IEEE Conference on Computer Vision and Pattern Recognition}, 2018, pp.
	8697--8710.
	
	\bibitem{real2017large}
	E.~Real, S.~Moore, A.~Selle, S.~Saxena, Y.~L. Suematsu, J.~Tan, Q.~Le, and
	A.~Kurakin, ``Large-scale evolution of image classifiers,'' in
	\emph{Proceedings of Machine Learning Research}, Sydney, Australia, 2017, pp.
	2902--2911.
	
	\bibitem{suganuma2017genetic}
	M.~Suganuma, S.~Shirakawa, and T.~Nagao, ``A genetic programming approach to
	designing convolutional neural network architectures,'' in \emph{Proceedings
		of the 2017 Genetic and Evolutionary Computation Conference}.\hskip 1em plus
	0.5em minus 0.4em\relax Berlin, Germany: ACM, 2017, pp. 497--504.
	
	\bibitem{zoph2016neural}
	B.~Zoph and Q.~V. Le, ``Neural architecture search with reinforcement
	learning,'' in \emph{Proceedings of the 2017 International Conference on
		Learning Representations}, Toulon, France, 2017.
	
	\bibitem{baker2016designing}
	B.~Baker, O.~Gupta, N.~Naik, and R.~Raskar, ``Designing neural network
	architectures using reinforcement learning,'' in \emph{Proceedings of the
		2017 International Conference on Learning Representations}, Toulon, France,
	2017.
	
	\bibitem{krizhevsky2009learning}
	A.~Krizhevsky and G.~Hinton, ``Learning multiple layers of features from tiny
	images,'' \emph{online: http://www.cs.toronto.edu/kriz/cifar.html}, 2009.
	
	\bibitem{back1996evolutionary}
	T.~Back, \emph{Evolutionary Algorithms in Theory and Practice: Evolution
		Strategies, Evolutionary Programming, Genetic Algorithms}.\hskip 1em plus
	0.5em minus 0.4em\relax England, UK: Oxford university press, 1996.
	
	\bibitem{sutton1998reinforcement}
	R.~S. Sutton and A.~G. Barto, \emph{Reinforcement learning: an
		introduction}.\hskip 1em plus 0.5em minus 0.4em\relax MIT press Cambridge,
	1998, vol.~1, no.~1.
	
	\bibitem{sun2017evolving}
	Y.~Sun, B.~Xue, and M.~Zhang, ``Evolving deep convolutional neural networks for
	image classification,'' \emph{arXiv preprint arXiv:1710.10741}, 2017.
	
	\bibitem{davis1991handbook}
	L.~Davis, \emph{Handbook of genetic algorithms}.\hskip 1em plus 0.5em minus
	0.4em\relax Bosa Roca, USA: Taylor \& Francis Inc, 1991.
	
	\bibitem{banzhaf1998genetic}
	W.~Banzhaf, P.~Nordin, R.~E. Keller, and F.~D. Francone, \emph{Genetic
		programming: an introduction}.\hskip 1em plus 0.5em minus 0.4em\relax Morgan
	Kaufmann San Francisco, 1998, vol.~1.
	
	\bibitem{janis1976evolutionary}
	C.~Janis, ``The evolutionary strategy of the equidae and the origins of rumen
	and cecal digestion,'' \emph{Evolution}, vol.~30, no.~4, pp. 757--774, 1976.
	
	\bibitem{schmitt2001theory}
	L.~M. Schmitt, ``Theory of genetic algorithms,'' \emph{Theoretical Computer
		Science}, vol. 259, no. 1-2, pp. 1--61, 2001.
	
	\bibitem{sun2018igd}
	Y.~Sun, G.~G. Yen, and Z.~Yi, ``{IGD} indicator-based evolutionary algorithm
	for many-objective optimization problems,'' \emph{IEEE Transactions on
		Evolutionary Computation}, vol.~23, no.~2, pp. 173--187, 2019.
	
	\bibitem{deb2002fast}
	K.~Deb, A.~Pratap, S.~Agarwal, and T.~Meyarivan, ``A fast and elitist
	multiobjective genetic algorithm: {NSGA-II},'' \emph{IEEE Transactions on
		Evolutionary Computation}, vol.~6, no.~2, pp. 182--197, 2002.
	
	\bibitem{sun2017reference}
	Y.~Sun, G.~G. Yen, and Z.~Yi, ``Reference line-based estimation of distribution
	algorithm for many-objective optimization,'' \emph{Knowledge-Based Systems},
	vol. 132, pp. 129--143, 2017.
	
	\bibitem{transferjiang2018}
	M.~Jiang, Z.~Huang, L.~Qiu, W.~Huang, and G.~G. Yen, ``Transfer learning based
	dynamic multiobjective optimization algorithms,'' \emph{IEEE Transactions on
		Evolutionary Computation}, vol.~22, no.~4, pp. 501--514, 2018.
	
	\bibitem{sun2018improve}
	Y.~Sun, G.~G. Yen, and Z.~Yi, ``Improved regularity model-based {EDA} for
	many-objective optimization,'' \emph{IEEE Transactions on Evolutionary
		Computation}, vol.~22, no.~5, pp. 662--678, 2018.
	
	\bibitem{mitchell1998introduction}
	M.~Mitchell, \emph{An introduction to genetic algorithms}.\hskip 1em plus 0.5em
	minus 0.4em\relax Cambridge, Massachusetts, USA: MIT press, 1998.
	
	\bibitem{LiuOn}
	H.~L. Liu, F.~Gu, Y.-m. Cheung, S.~Xie, and J.~Zhang, ``On solving {WCDMA}
	network planning using iterative power control scheme and evolutionary
	multiobjective algorithm [application notes],'' \emph{IEEE Computational
		Intelligence Magazine}, vol.~9, no.~1, pp. 44--52, 2014.
	
	\bibitem{Nag2015A}
	K.~Nag and N.~R. Pal, ``A multiobjective genetic programming-based ensemble for
	simultaneous feature selection and classification,'' \emph{IEEE Transactions
		on Cybernetics}, vol.~46, no.~2, pp. 499--510, 2015.
	
	\bibitem{hochreiter1997long}
	S.~Hochreiter and J.~Schmidhuber, ``Long short-term memory,'' \emph{Neural
		Computation}, vol.~9, no.~8, pp. 1735--1780, 1997.
	
	\bibitem{gers1999learning}
	F.~A. Gers, J.~Schmidhuber, and F.~Cummins, ``Learning to forget: continual
	prediction with lstm,'' \emph{Neural Computation}, vol.~12, no.~10, pp.
	2451--2471, 2000.
	
	\bibitem{srivastava2015training}
	R.~K. Srivastava, K.~Greff, and J.~Schmidhuber, ``Training very deep
	networks,'' in \emph{Advances in Neural Information Processing Systems},
	Montréal, Canada, 2015, pp. 2377--2385.
	
	\bibitem{orhan2017skip}
	A.~E. Orhan and X.~Pitkow, ``Skip connections eliminate singularities,'' in
	\emph{Proceedings of 2018 Machine Learning Research}, Stockholm, Sweden,
	2018.
	
	\bibitem{srinivas1994genetic}
	M.~Srinivas and L.~M. Patnaik, ``Genetic algorithms: A survey,''
	\emph{Computer}, vol.~27, no.~6, pp. 17--26, 1994.
	
	\bibitem{hawkins2004problem}
	D.~M. Hawkins, ``The problem of overfitting,'' \emph{Journal of Chemical
		Information and Computer Sciences}, vol.~44, no.~1, pp. 1--12, 2004.
	
	\bibitem{srivastava2014dropout}
	N.~Srivastava, G.~Hinton, A.~Krizhevsky, I.~Sutskever, and R.~Salakhutdinov,
	``Dropout: a simple way to prevent neural networks from overfitting,''
	\emph{The Journal of Machine Learning Research}, vol.~15, no.~1, pp.
	1929--1958, 2014.
	
	\bibitem{he2016identity}
	K.~He, X.~Zhang, S.~Ren, and J.~Sun, ``Identity mappings in deep residual
	networks,'' in \emph{Lecture Notes in Computer Science}.\hskip 1em plus 0.5em
	minus 0.4em\relax Amsterdam, the Netherlands: Springer, 2016, pp. 630--645.
	
	\bibitem{housley2004224}
	R.~Housley, ``A 224-bit one-way hash function: Sha-224,'' RFC 3874, September
	2004.
	
	\bibitem{nasrabadi2007pattern}
	N.~M. Nasrabadi, ``Pattern recognition and machine learning,'' \emph{Journal of
		Electronic Imaging}, vol.~16, no.~4, p. 049901, 2007.
	
	\bibitem{glorot2011deep}
	X.~Glorot, A.~Bordes, and Y.~Bengio, ``Deep sparse rectifier neural networks,''
	in \emph{Proceedings of the 14th International Conference on Artificial
		Intelligence and Statistics}, FL, USA, 2011, pp. 315--323.
	
	\bibitem{NIPS2017_6790}
	S.~Ioffe, ``Batch renormalization: towards reducing minibatch dependence in
	match-normalized models,'' in \emph{Advances in Neural Information Processing
		Systems}.\hskip 1em plus 0.5em minus 0.4em\relax Curran Associates, Inc.,
	2017, pp. 1945--1953.
	
	\bibitem{bottou2012stochastic}
	L.~Bottou, ``Stochastic gradient descent tricks,'' in \emph{Neural networks:
		tricks of the trade}.\hskip 1em plus 0.5em minus 0.4em\relax Springer, 2012,
	pp. 421--436.
	
	\bibitem{helfenstein2012parallel}
	R.~Helfenstein and J.~Koko, ``Parallel preconditioned conjugate gradient
	algorithm on gpu,'' \emph{Journal of Computational and Applied Mathematics},
	vol. 236, no.~15, pp. 3584--3590, 2012.
	
	\bibitem{abadi2016tensorflow}
	M.~Abadi, P.~Barham, J.~Chen, Z.~Chen, A.~Davis, J.~Dean, M.~Devin,
	S.~Ghemawat, G.~Irving, M.~Isard \emph{et~al.}, ``Tensorflow: a system for
	large-scale machine learning.'' in \emph{Operating Systems Design and
		Implementation}, vol.~16, 2016, pp. 265--283.
	
	\bibitem{paszke2017automatic}
	\BIBentryALTinterwordspacing
	A.~Paszke, S.~Gross, S.~Chintala, G.~Chanan, E.~Yang, Z.~DeVito, Z.~Lin,
	A.~Desmaison, L.~Antiga, and A.~Lerer, ``Automatic differentiation in
	pytorch,'' 2017. [Online]. Available:
	\url{https://openreview.net/forum?id=BJJsrmfCZ}
	\BIBentrySTDinterwordspacing
	
	\bibitem{miller1995genetic}
	B.~L. Miller, D.~E. Goldberg \emph{et~al.}, ``Genetic algorithms, tournament
	selection, and the effects of noise,'' \emph{Complex Systems}, vol.~9, no.~3,
	pp. 193--212, 1995.
	
	\bibitem{michalewicz1996genetic}
	Z.~Michalewicz and S.~J. Hartley, ``Genetic algorithms + data structures =
	evolution programs,'' \emph{Mathematical Intelligencer}, vol.~18, no.~3,
	p.~71, 1996.
	
	\bibitem{goldberg1988genetic}
	D.~E. Goldberg and J.~H. Holland, ``Genetic algorithms and machine learning,''
	\emph{Machine Learning}, vol.~3, no.~2, pp. 95--99, 1988.
	
	\bibitem{anderson2005practical}
	C.~M. Anderson-Cook, ``Practical genetic algorithms,'' p. 1099, 2005.
	
	\bibitem{malik2014preventing}
	S.~Malik and S.~Wadhwa, ``Preventing premature convergence in genetic algorithm
	using dgca and elitist technique,'' \emph{nternational Journal of Advanced
		Research in Computer Science and Software Engineering}, vol.~4, no.~6, 2014.
	
	\bibitem{zhang2003novel}
	G.~Zhang, Y.~Gu, L.~Hu, and W.~Jin, ``A novel genetic algorithm and its
	application to digital filter design,'' in \emph{Proceedings of 2003 IEEE
		Intelligent Transportation Systems}, vol.~2.\hskip 1em plus 0.5em minus
	0.4em\relax IEEE, 2003, pp. 1600--1605.
	
	\bibitem{bhandari1996genetic}
	D.~Bhandari, C.~Murthy, and S.~K. Pal, ``Genetic algorithm with elitist model
	and its convergence,'' \emph{International Journal of Pattern Recognition and
		Artificial Intelligence}, vol.~10, no.~06, pp. 731--747, 1996.
	
	\bibitem{xie2009sampling}
	H.~Xie and M.~Zhang, ``Impacts of sampling strategies in tournament selection
	for genetic programming,'' \emph{Soft Computing}, vol.~16, no.~4, pp.
	615--633, 2012.
	
	\bibitem{goodfellow2013maxout}
	I.~J. Goodfellow, D.~Warde-Farley, M.~Mirza, A.~Courville, and Y.~Bengio,
	``Maxout networks,'' in \emph{Proceedings of the 30th International
		Conference on Machine Learning}, Atlanta, Georgia, USA, Jun 2013, pp.
	1319--1327.
	
	\bibitem{lin2013network}
	M.~Lin, Q.~Chen, and S.~Yan, ``Network in network,'' in \emph{Proceedings of
		the 2014 International Conference on Learning Representations}, Banff,
	Canada, 2014.
	
	\bibitem{srivastava2015highway}
	R.~K. Srivastava, K.~Greff, and J.~Schmidhuber, ``Highway networks,'' in
	\emph{Proceedings of the 2015 International Conference on Learning
		Representations Workshop}, San Diego, CA, 2015.
	
	\bibitem{springenberg2014striving}
	J.~T. Springenberg, A.~Dosovitskiy, T.~Brox, and M.~Riedmiller, ``Striving for
	simplicity: the all convolutional net,'' in \emph{Proceedings of the 2015
		International Conference on Learning Representations}, San Diego, CA, 2015.
	
	\bibitem{ILSVRC15}
	O.~Russakovsky, J.~Deng, H.~Su, J.~Krause, S.~Satheesh, S.~Ma, Z.~Huang,
	A.~Karpathy, A.~Khosla, M.~Bernstein, A.~C. Berg, and L.~Fei-Fei, ``Imagenet
	large scale visual recognition challenge,'' \emph{International Journal of
		Computer Vision}, vol. 115, no.~3, pp. 211--252, 2015.
	
	\bibitem{liu2018darts}
	H.~Liu, K.~Simonyan, and Y.~Yang, ``Darts: Differentiable architecture
	search,'' \emph{arXiv preprint arXiv:1806.09055}, 2018.
	
	\bibitem{devries2017improved}
	T.~DeVries and G.~W. Taylor, ``Improved regularization of convolutional neural
	networks with cutout,'' \emph{arXiv preprint arXiv:1708.04552}, 2017.
	
	\bibitem{wolpert1997no}
	D.~H. Wolpert and W.~G. Macready, ``No free lunch theorems for optimization,''
	\emph{IEEE Transactions on Evolutionary Computation}, vol.~1, no.~1, pp.
	67--82, 1997.
	
	\bibitem{dalal2005histograms}
	N.~Dalal and B.~Triggs, ``Histograms of oriented gradients for human
	detection,'' in \emph{Proceedings of the IEEE Conference on Computer Vision
		and Pattern Recognition}, 2005.
	
	\bibitem{williamson1989box}
	D.~F. Williamson, R.~A. Parker, and J.~S. Kendrick, ``The box plot: a simple
	visual method to interpret data,'' \emph{Annals of Internal Medicine}, vol.
	110, no.~11, pp. 916--921, 1989.
	
\end{thebibliography}
\end{document}